\documentclass{article} 
\usepackage{iclr2025_conference,times}

\usepackage{amsmath,amsfonts,bm}

\def\eqref#1{equation~\ref{#1}}

\def\1{\bm{1}}

\def\vs{{\bm{s}}}

\DeclareMathAlphabet{\mathsfit}{\encodingdefault}{\sfdefault}{m}{sl}
\SetMathAlphabet{\mathsfit}{bold}{\encodingdefault}{\sfdefault}{bx}{n}

\usepackage{xcolor}
\usepackage{multirow}
\usepackage{hyperref}
\usepackage{url}
\usepackage{booktabs}
\usepackage{graphicx}
\usepackage{wrapfig}
\usepackage{subcaption}
\usepackage{enumitem}
\usepackage{tcolorbox}
\usepackage[capitalise,noabbrev]{cleveref}

\newlist{inparaenum}{enumerate*}{1}
\setlist[inparaenum,1]{label=(\arabic*), itemjoin={{ }}, itemjoin*={{ }}}

\def\eg{{\em e.g.,}\xspace}
\def\vs{{\em v.s.}\xspace}
\def\ie{{\em i.e.,}\xspace}

\definecolor{darkgreen}{RGB}{0,100,0}
\definecolor{darkred}{RGB}{139,0,0}

\title{Lost in Time: Clock and Calendar Understanding Challenges in Multimodal LLMs}

\author{Rohit Saxena\textsuperscript{\includegraphics[width=.025\textwidth]{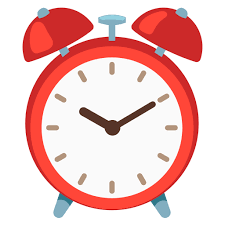}} \quad Aryo Pradipta Gema\textsuperscript{\includegraphics[width=.025\textwidth]{./figures/clock}} \quad Pasquale Minervini\textsuperscript{\includegraphics[width=.025\textwidth]{./figures/clock} \includegraphics[width=.025\textwidth]{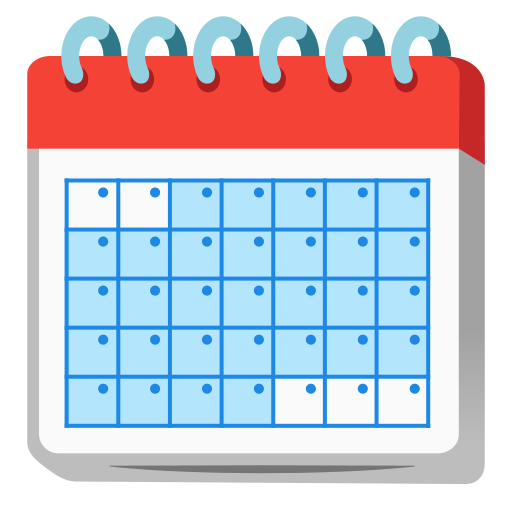}} \\
\textsuperscript{\includegraphics[width=.025\textwidth]{./figures/clock}}ILCC, School of Informatics, University of Edinburgh \qquad 
\textsuperscript{\includegraphics[width=.025\textwidth]{./figures/calendar}}Miniml.AI\\
\texttt{\{rohit.saxena, aryo.gema, p.minervini\}@ed.ac.uk} \\
}

\usepackage{xspace}
\newcommand{\dataset}{DateTimeReasoning\xspace}

\iclrfinalcopy 
\begin{document}
\maketitle
\begin{abstract}
Understanding time from visual representations is a fundamental cognitive skill, yet it remains a challenge for multimodal large language models (MLLMs).
In this work, we investigate the capabilities of MLLMs in interpreting time and date through analogue clocks and yearly calendars.
To facilitate this, we curated a structured dataset\footnote{\href{https://huggingface.co/datasets/rohitsaxena/DateTimeQA}{https://huggingface.co/datasets/rohitsaxena/DateTimeQA}} comprising two subsets:
\begin{inparaenum}
\item \emph{ClockQA}, which comprises various types of clock styles—standard, black-dial, no-second-hand, Roman numeral, and arrow-hand clocks—paired with time-related questions; and
\item \emph{CalendarQA}, which consists of yearly calendar images with questions ranging from commonly known dates (\eg Christmas, New Year’s Day) to computationally derived ones (\eg the 100th or 153rd day of the year).
\end{inparaenum}
We aim to analyse how MLLMs can perform visual recognition, numerical reasoning, and temporal inference when presented with time-related visual data.
Our evaluations show that despite recent advancements, reliably understanding time remains a significant challenge for MLLMs.
\end{abstract}
\section{Introduction}
\begin{wrapfigure}{r}{0.58\textwidth}
    \centering
    \vspace{-15pt}
    \includegraphics[width=0.58\textwidth]{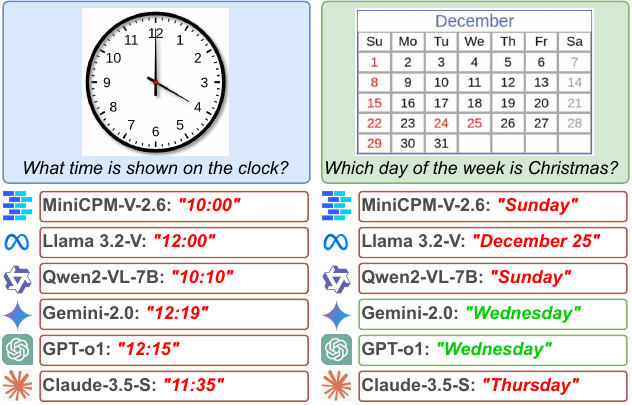}
    \caption{Predictions on ClockQA and CalendarQA.}
    \label{fig:error_examples}
    \vspace{-5pt}
\end{wrapfigure}
The ability to interpret and reason about time from visual inputs is critical for many real‐world applications—ranging from event scheduling to autonomous systems.
Despite advances in multimodal large language models (MLLMs), most work has focused on object detection~\citep{DetToolChain}, image captioning~\citep{MM1}, or scene understanding~\citep{fu2024mmecomprehensiveevaluationbenchmark}, leaving temporal inference underexplored~\citep{zhang2025mmerealworld}.
In particular, analogue clock reading and calendar comprehension involve intricate cognitive steps: they demand fine‐grained visual recognition (\eg clock‐hand position, day‐cell layout) and non-trivial numerical reasoning (\eg calculating day offsets).
In recent years, a variety of vision-language benchmarks were proposed to evaluate multimodal reasoning on diverse tasks such as geometry, logic, coding, and advanced mathematics~\citep{Yue_2024_CVPR, kazemi2024geomverse, kazemi2024remi, lu2024mathvista}.
Additional efforts have been made to automatically read analogue clocks and other dials~\citep{yang2022its, Alexeev20, bao19, cai20, Howells_2021_CVPR}, showing that dial or gauge interpretation is a cognitively complex skill requiring visual-spatial understanding and arithmetic reasoning.
However, clock and calendar readings remain underexplored in these existing large-scale benchmarks, and comprehensive evaluations of MLLMs on such tasks are lacking (see \cref{sec:related} for more information on related works).

In this paper, we explore how MLLMs handle these temporal tasks.
We constructed a focused test set consisting of two subsets: \emph{ClockQA}, which includes diverse analogue clock images across six categories (including variations with Roman numerals, missing second hands, and different dial colours) paired with time‐related questions, and \emph{CalendarQA}, which comprises 10 years of calendar images paired with questions ranging from straightforward date lookups ( What day of the week is New Year's Day?) to more complex queries (What is the 153rd day of the year?).
While our dataset is intentionally small in scale, it is designed to probe specific aspects of temporal reasoning, visual parsing, and date/time inference.

We evaluate multiple state-of-the-art closed and open-source models on these tasks.
Our preliminary findings reveal that while some models show promise in clock reading (\eg Gemini‐2.0 demonstrates lower hour and minute errors) or in calendar question‐answering (\eg o1 exhibits high accuracy), persistent challenges remain.
Despite the limited scale of our evaluation, error analyses highlight challenges in correctly parsing clock‐hand positions and performing arithmetic on dates for calendar tasks.
These insights provide valuable direction for future work in the temporal reasoning capabilities of MLLMs.
\begin{figure}[t]
    \centering\scalebox{0.8}{%
    \includegraphics[width=\textwidth]{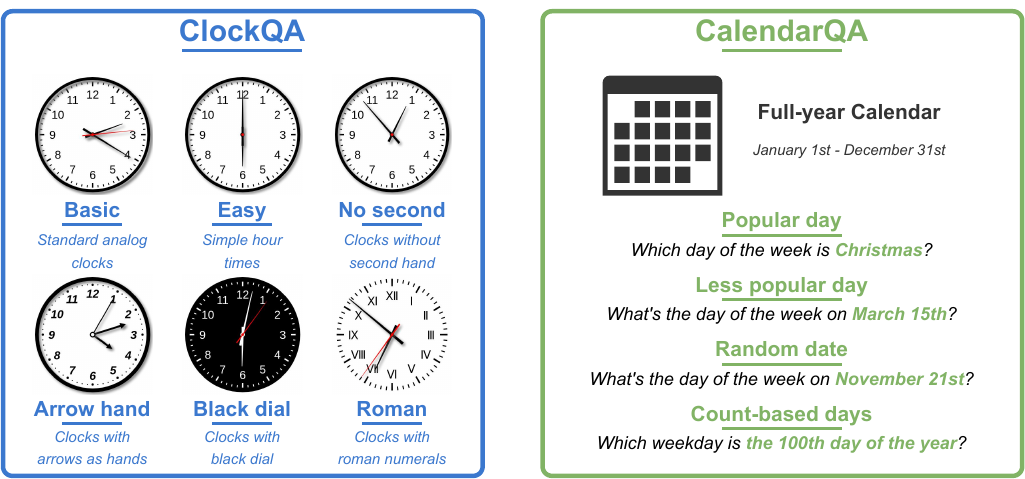}}
    \caption{Overview of \dataset and its two main subsets: ClockQA and CalendarQA}
    \label{fig:dataset}
\end{figure}
\begin{table}[t]
    \centering
    \scalebox{0.80}{%
    \begin{tabular}{lcccc| cccc}
        \toprule
        \multirow{2}{*}{\textbf{Model}} & \multicolumn{4}{c|}{\textbf{Clock}} & \multicolumn{4}{c}{\textbf{Calendar}} \\
        \cmidrule(lr){2-5}
        \cmidrule(lr){6-9}
        & \textbf{EM}$\uparrow$ & \textbf{MAE}$\downarrow$ & \textbf{Hour Err}$\downarrow$ & \textbf{Min Err}$\downarrow$ & \textbf{Acc}$\uparrow$ & \textbf{P}$\uparrow$ & \textbf{R}$\uparrow$ & \textbf{F1}$\uparrow$ 
        \\
        \midrule
        Llama 3.2-Vision & 3.23 & 10825.84 & 3.02 & 13.08 & 11.67 & 9.18 & 11.67 & 10.03\\
        Qwen2-VL-7B & 0.0 & 11167.71 & 3.06 & 13.16 & 18.33 & 9.51 & 18.33 & 12.09\\
        MiniCPM-V-2.6 & 3.23 & 11078.79 & 3.06 & 10.73 & 20.0 & 19.18 & 20.0 & 15.61\\
        Gemini-2.0 & \textbf{22.58} & \textbf{6494.37} & \textbf{1.82} & \textbf{6.4} &  31.67 & 30.32 & 31.67 & 29.79\\
        GPT-4o & 8.06 & 8268.42 & 2.29 & 11.97 & 43.33 & 46.12 & 43.33 & 42.12\\
        Claude-3-5-sonnet & 6.45 & 7964.24 & 2.13 & 12.32 & 46.67 & 42.56 & 46.67 & 43.79\\
        GPT-o1 & 4.84 & 7954.45 & 2.21 & 8.19 & \textbf{80.0} & \textbf{80.39} & \textbf{80.0} & \textbf{80.04}\\
        \bottomrule
    \end{tabular}
    }
    \caption{Performance of each model on Clock (left) and Calendar (right) tasks. 
    Higher values are better ($\uparrow$); lower values are better ($\downarrow$).}
    \label{tab:results}
\end{table}
\section{Dataset}\label{sec:dataset}
We created a small dataset comprising two subsets, \emph{ClockQA} and \emph{CalendarQA}, each containing images paired with question-answer pairs to test the time and date reasoning of multimodal large language models (MLLMs). Figure~\ref{fig:dataset} illustrates the dataset and its two subsets.
\textbf{ClockQA.}
Given an image of an analogue clock, a multimodal LLM is asked the following question \emph{``What time is shown on the clock in the given image?''}
This requires (1) detecting the clock hand positions (hour, minute, and second) and (2) converting them into time representation.
The ClockQA subset contains 62 samples of analogue clocks with varying appearances, requiring precise readings of the hour, minute, and second hands. It includes the following categories:
\begin{inparaenum}
    \item \textbf{Basic clocks:} standard analogue clocks;
    \item \textbf{Black dial clocks:} featuring a darker face for contrast-based parsing;
    \item \textbf{No second hand clocks:} simplified version of the task;
    \item \textbf{Easy clocks:} standard clock set on the hour (\eg 4:00);
    \item \textbf{Roman number clocks:} for digit-recognition challenges; and
    \item \textbf{Arrow hand clocks:} stylized with arrows as the hands for more obvious pointer cues.
\end{inparaenum}
\textbf{CalendarQA.}
Given a full image of a yearly calendar (January 1--December 31), the model answers questions from one of the categories like \emph{“Which day of the week is New Year’s Day?”} or \emph{“What is the 153rd day of the year?”}. The system must combine visual parsing of date cells and textual labels with date arithmetic or reasoning about day offsets.
The CalendarQA subset spans 10 full years (January 1 to December 31), each with six questions focusing on four main categories:
\begin{inparaenum}
    \item \textbf{Popular days} (\eg Christmas and New Year’s);
    \item \textbf{Less popular dates} such as the Ides of March (March 15);
    \item \textbf{Random dates} like November 21; and
    \item \textbf{Count-based days} referring to the $n$th day of the year (\eg the 100th).
\end{inparaenum}

Together, these two subsets evaluate an MLLM’s ability to parse and reason about time and date information in a multimodal context.
\section{Tasks and Experiments}
\label{sec:tasks_experiments}
\textbf{Experimental Setup.}
We evaluate seven multimodal LLMs in a zero-shot setting.
We evaluate closed-source multimodal models, including GPT-4o~\citep{openai2024gpt4technicalreport}, GPT-o1~\citep{openai2024gpt4technicalreport}, Gemini 2.0~\citep{geminiteam2024}, and  Claude 3.5 Sonnet~\citep{anthropic_claude3_5_sonnet}.
We also evaluate open-source models such as Llama 3.2-11B-Vision-Instruct~\citep{lama3herdmodels}, Qwen2-VL-7B-Instruct~\citep{qwen2technicalreport}, and MiniCPM-V-2.6~\citep{yao2024minicpm}.
The experiment details and exact prompt template used are in Appendix~\ref{sec:exp_prompt_template}.
\textbf{ClockQA Metrics.}
We measure performance using four metrics.
\textbf{Exact Match (EM)} is the proportion of predicted clock readings that exactly match the ground truth.
\textbf{MAE (Seconds)} quantifies the average absolute difference between predicted and actual times in seconds, applying a circular 12-hour wraparound (maximum error: 21,600 seconds).
We also report \textbf{Hour Error and Minute Error}, which compute mean absolute differences for each clock hand, again with a circular wraparound.
See \cref{app:metrics_clockqa} for more details.

\textbf{CalendarQA Metrics.}
We adopt standard classification metrics:
\textbf{Accuracy (Acc)} to measure correct weekday predictions, and macro-averaged \textbf{Precision (P)}, \textbf{Recall (R)}, and \textbf{F1} across date categories.
See \cref{app:metrics_calendarqa} for more details.

\textbf{Implementation Details.}
We conduct experiments on a shared test set of 62 clock samples (across six clock‐face variants) and 10 calendar years (with six question types per year).
Model prompts followed a consistent format, providing one image and one question.
Responses are automatically parsed to extract time or weekday (\ie removing explanation or conversion of short forms) and evaluated against the reference answer. 
\begin{figure}[t]
    \centering
    \begin{subfigure}{0.49\textwidth}
    \includegraphics[width=\textwidth]{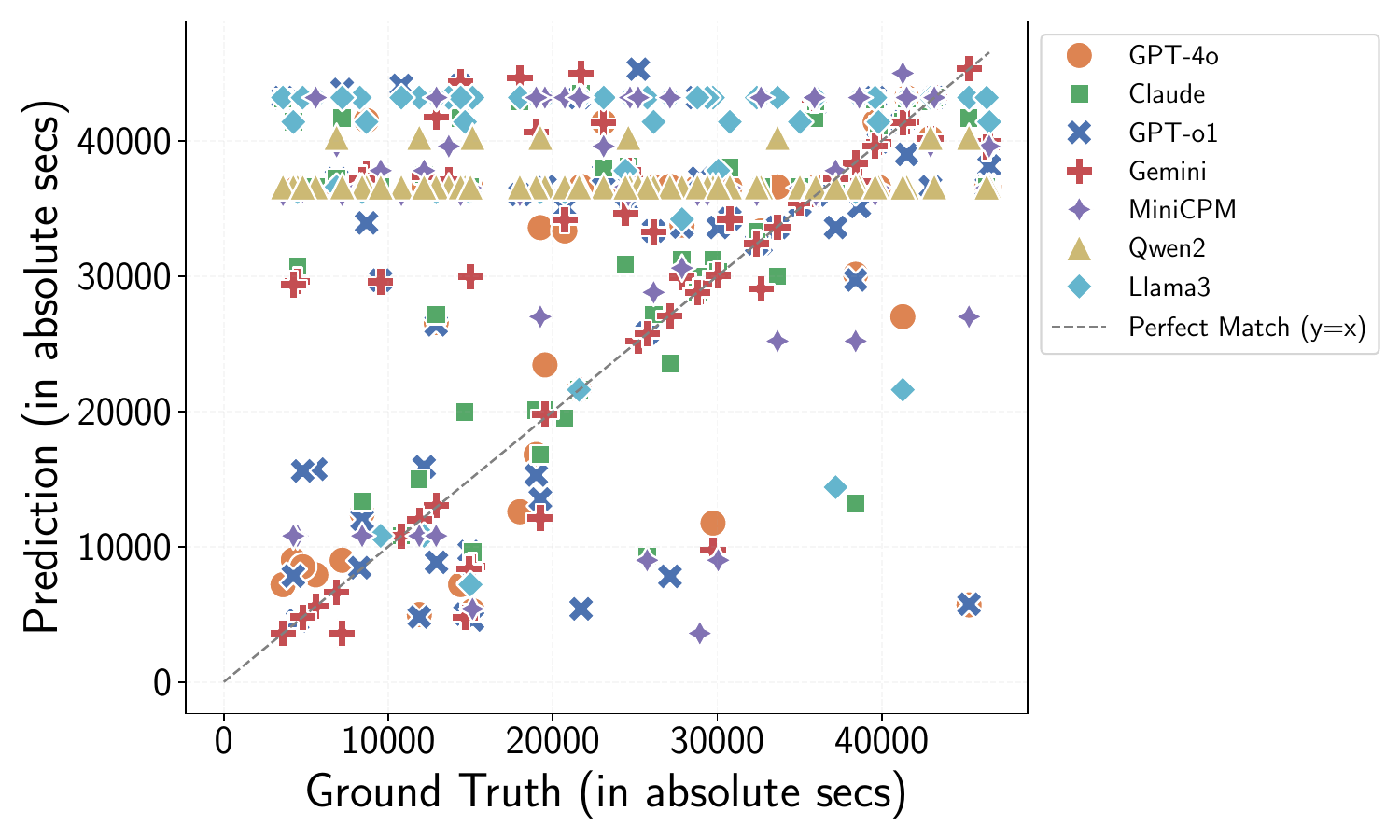}
    \caption{Points represent predicted times (s) by models \vs ground truth (x-axis). The dashed black line (y = x) represents a perfect model. Models show varying errors from this line.}
    \label{fig:clock_absolute_error}
     \end{subfigure}  
     \begin{subfigure}{0.48\textwidth}
    \includegraphics[width=\textwidth]{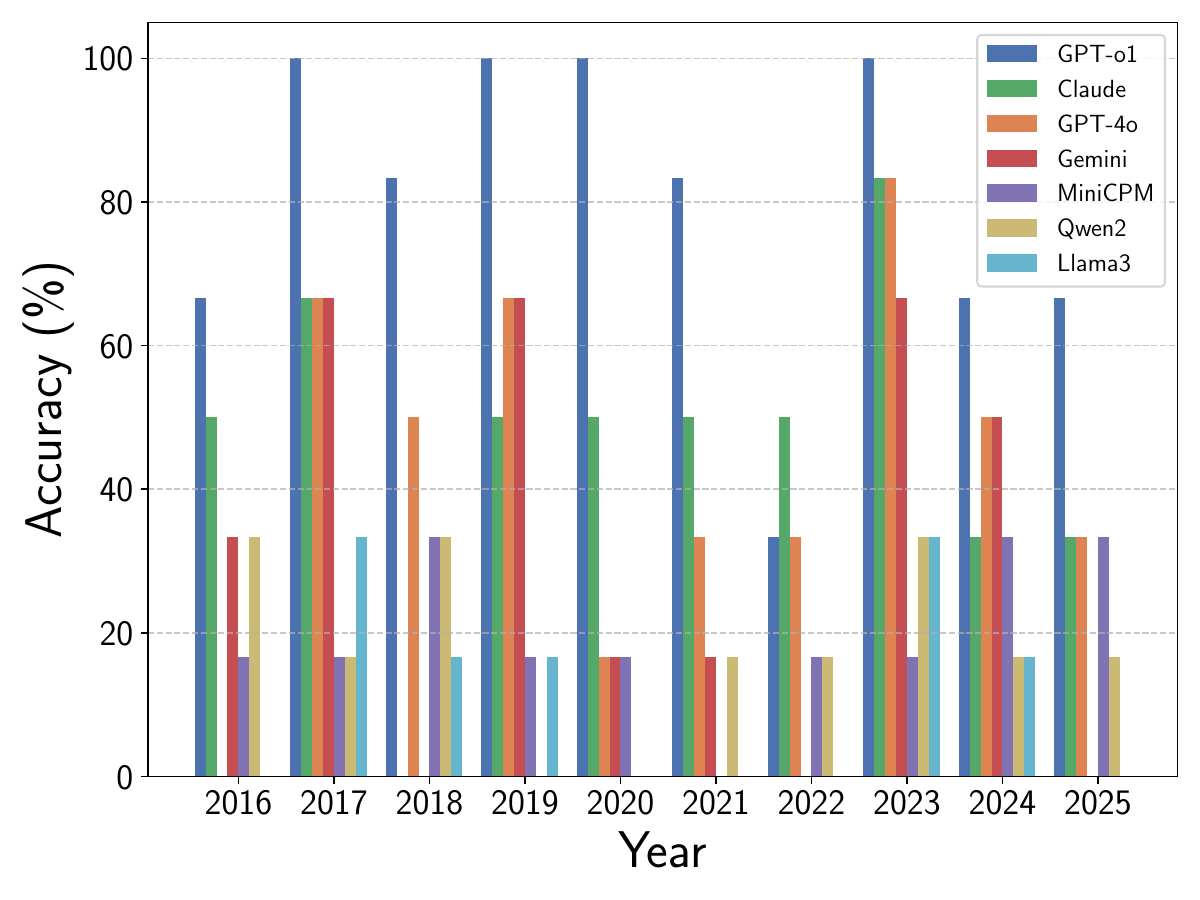}
    \caption{Year-wise accuracy of the models. Blank bar indicates accuracy as $0\%$ for that year.}
    \label{fig:calendar_year_wise}
     \end{subfigure}  
     \caption{Error analysis for ClockQA and CalendarQA.}
\end{figure}
\begin{figure}[t]
\centering
    \begin{subfigure}{0.48\textwidth}
        \includegraphics[width=\textwidth]{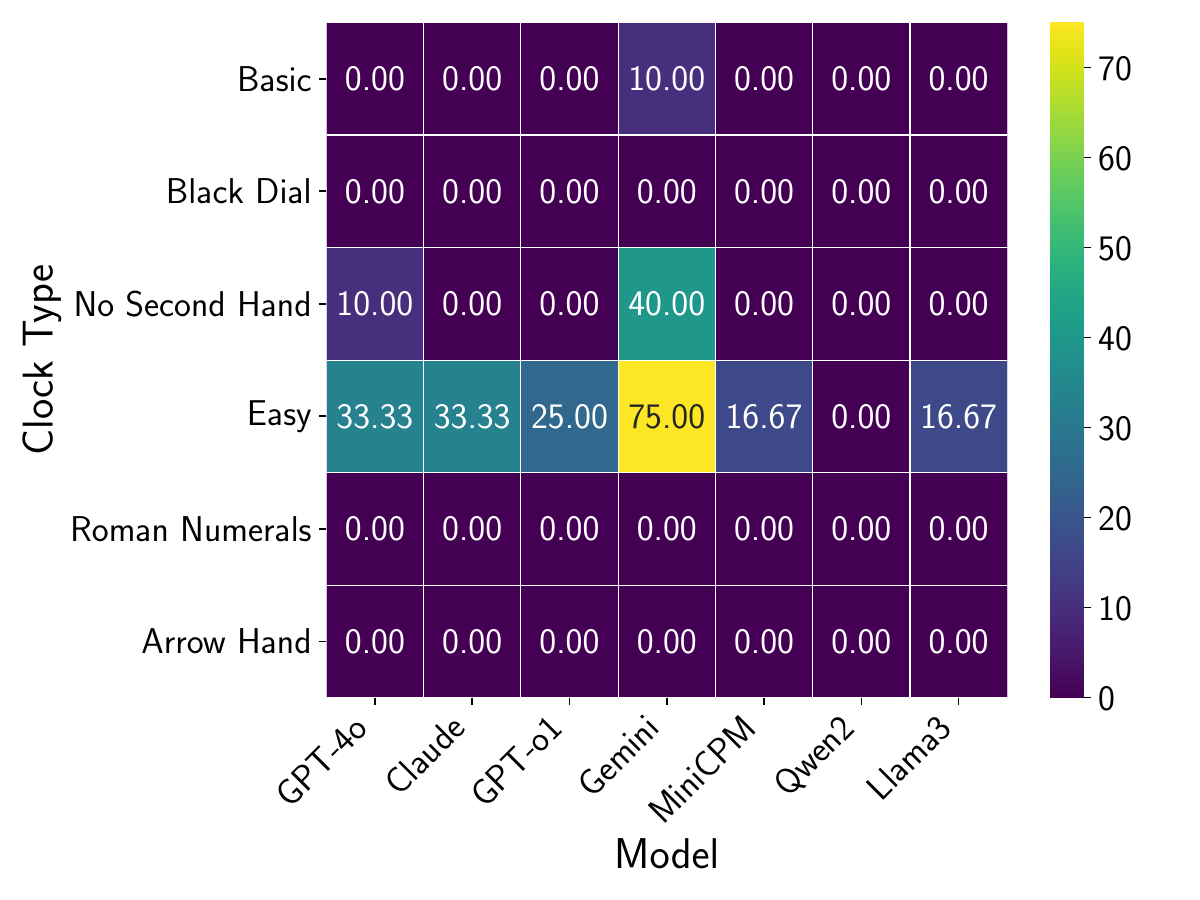}
        \caption{Clock category-wise accuracy of the models.}
        \label{fig:clock_type_EM_heatmap}
    \end{subfigure}
    \begin{subfigure}{0.49\textwidth}
        \includegraphics[width=\textwidth]{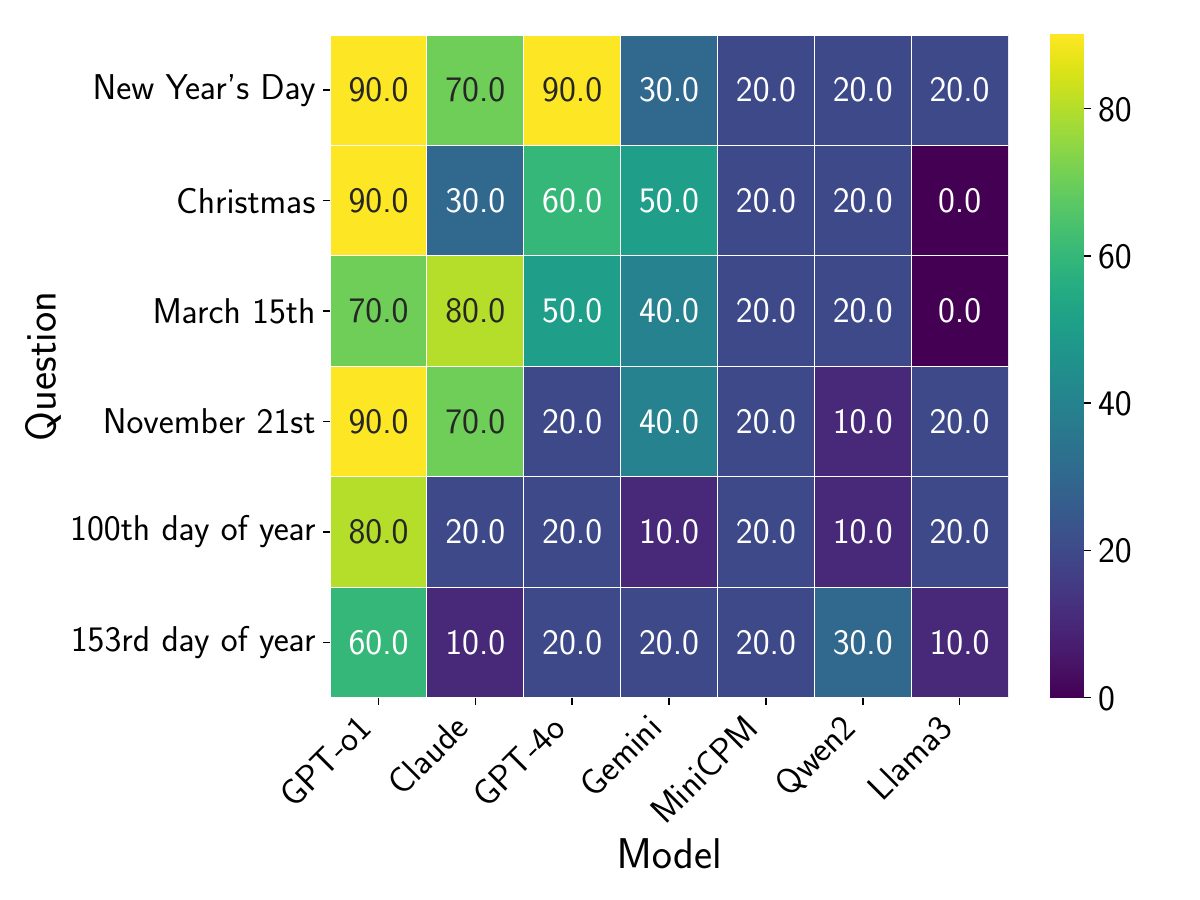}
        \caption{Calendar question-wise accuracy of the models.}
        \label{fig:day_of_week_correctness_heatmap}
    \end{subfigure}
    \caption{ClockQA and CalendarQA question and category-based analysis.}
\end{figure}
\section{Results and Discussion}
\label{sec:result}
\cref{tab:results} summarizes performance across both tasks. In ClockQA, \textbf{Gemini-2.0} achieves the highest EM score (22.58\%) and the lowest hour/minute errors, indicating relatively stronger clock understanding compared to other models. However, overall EM scores remain low, underscoring persistent difficulties in clock reading by MLLMS. Conversely, \textbf{GPT-o1} excels in CalendarQA with an accuracy of 80\%, highlighting robust date arithmetic and reasoning capabilities. Other models lag substantially, indicating that date arithmetic and structured layout parsing remain challenging.  Overall performance on both ClockQA and CalendarQA remains poor, except for the high performance of GPT-o1 on CalendarQA. See Appendix \ref{sec:prediction_samples} for a sample of generated predictions.

\textbf{Clock Reading Remains Error-Prone.} Across the ClockQA subset, performance was notably weaker than for the calendar questions (see \cref{tab:results}). Figures~\ref{fig:clock_type_EM_heatmap} and \ref{fig:clock_absolute_error} reveal that performance remains poor even on standard dials; some models exhibit bias toward a single “default” time. Roman numerals and stylized clock hands further increase the errors. Removing the second hand did not simplify reasoning, suggesting deep-seated issues with hand detection and angle interpretation.

\textbf{Calendar Reasoning Analysis} By contrast, calendar tasks elicited higher success rates for certain models and question types. GPT-o1 dominates the CalendarQA subset with an 80\% overall accuracy (\cref{tab:results} and \cref{fig:calendar_year_wise}).

Closed-source models like GPT-o1 and Claude-3.5 outshine open-source ones on popular holidays, potentially reflecting memorized patterns in the training data (see \cref{fig:day_of_week_correctness_heatmap}). However, accuracy diminishes substantially for lesser-known or arithmetically demanding queries (\eg 153rd day), indicating that performance does not transfer well to offset-based reasoning. The drop is especially evident among smaller or open-source models (MiniCPM, Qwen2-VL-7B, and Llama3.2-Vision), which exhibit near-random performance on less popular or offset-based queries.
\section{Conclusion}
In this work, we conduct a preliminary study on understanding and reasoning about time from visual inputs, which remains a significant challenge for multimodal large language models. 
We build a small dataset to benchmark these models for clock and calendar understanding.
The experimental results highlight key shortcomings in the ability of these models to accurately interpret time from analogue clocks and yearly calendars. 
Our findings suggest that successful temporal reasoning requires a combination of precise visual perception, numerical computation, and structured logical inference that current MLLMs have not yet mastered.
This work highlights the need for further research to improve the processing of geometric relationships in clock faces and structured calendar information in MLLMs.
\section*{Acknowledgments}
This work was supported in part by the School of Informatics at the University of Edinburgh.
Rohit Saxena was supported by the EPSRC (grant no.\ EP/V025708/1).
Aryo Pradipta Gema was supported by the United Kingdom Research and Innovation (grant EP/S02431X/1), UKRI Centre for Doctoral Training in Biomedical AI at the University of Edinburgh, School of Informatics.
Pasquale Minervini was partially funded by ELIAI (The Edinburgh Laboratory for Integrated Artificial Intelligence), EPSRC (grant no.\ EP/W002876/1), and a donation from Accenture LLP.
This work was also supported by the Edinburgh International Data Facility (EIDF) and the Data-Driven Innovation Programme at the University of Edinburgh.
\bibliography{iclr2025_conference}
\bibliographystyle{iclr2025_conference}
\clearpage
\appendix
\section{Related Work} \label{sec:related}
\textbf{Vision-Language Benchmarks.} Existing benchmarks for MLLMs cover various tasks, from college-level subject knowledge to advanced mathematical and multi-step reasoning.
Massive Multi-discipline Multimodal Understanding~\cite[MMMU, ][]{Yue_2024_CVPR} tests deliberate reasoning across 11.5K multimodal questions drawn from college exams, quizzes, and textbooks, spanning disciplines such as Art, Business, Science, Health, Social Science, and Engineering.
MathVista~\cite{lu2024mathvista} gauges mathematical reasoning within visual contexts, while GeomVerse~\cite{kazemi2024geomverse} evaluates geometry-based problem-solving in MLLMs.
ReMI~\cite{kazemi2024remi} focuses on multi-image reasoning, encompassing diverse domains, including math, physics, logic, code, table/chart comprehension, and spatio-temporal tasks.
Despite this breadth, none of these datasets specifically targets analogue clock and calendar interpretation.

\textbf{Analogue Clock and Dial Reading.}
Reading analogue clocks is a complex cognitive task that engages multiple mental processes~\cite{DBLP:journals/corr/abs-2410-11756}.
It involves several key cognitive components: visuospatial skills for understanding spatial relationships between clock elements, executive functioning for planning and reasoning, working memory to maintain mental representations of time concepts, and sustained attention to process the information accurately~\cite{DBLP:journals/corr/abs-2410-11756,5673a1d18644461d895f9b7c6ecfe2e1}.
\citet{yang2022its} provide a comprehensive framework for reading analogue clocks in natural images and videos, introducing a synthetic data generation pipeline which can produce a wide variety of clock images reflecting the challenges encountered in real-world scenes.
Recently, \citet{deitke2024molmopixmoopenweights} introduced a large-scale synthetic dataset of analogue clocks comprising images rendered from different watch models.  However, the dataset's generation pipeline is not publicly available, making it difficult to reproduce.
\citet{ghosal2024languagemodelspuzzleprodigies} propose a puzzle-based task that includes clock puzzles to identify when an event occurred or will occur, given a starting time and an elapsed or future duration. 
Another line of work focuses on automatic dial or gauge meter readings.
The solutions proposed for dial reading rely on neural models~\cite{Alexeev20}, projective transforms~\cite{bao19}, and virtual dataset generators~\cite{cai20, Howells_2021_CVPR}, which produce accurate results for gauges with known shape and style.
\section{Analysis of Easy Clock Predictions}
\begin{table}[ht]
  \centering
  \begin{tabular}{ccl}
    \toprule
    \textbf{Time} & \textbf{Number of Models} & \textbf{Models with wrong prediction} \\
    \midrule
    1:00  & 6  & All except Gemini\\
    2:00  & 7 & All \\ 
    3:00  & 4 & Llama3, Qwen2, MiniCPM, GPT-o1\\ 
    4:00  & 7 & All \\
    5:00  & 7 & All \\
    6:00  & 3 & Qwen2, MiniCPM, GPT-o1\\
    7:00  & 6 & All except Gemini\\
    8:00  & 5 & Llama3, Qwen2, MiniCPM, GPT-4o, GPT-o1\\
    9:00  & 4 & Llama3, Qwen2, MiniCPM, Claude, GPT-o1\\
    10:00 & 4 & Llama3, Qwen2, Claude, GPT-4o\\
    11:00 & 6 & All except Gemini\\
    12:00 & 1 & Qwen2\\
    \bottomrule
  \end{tabular}
    \caption{Number of models with incorrect predictions at each hour (12-hour format).}
    \label{tab:error_frequency_models}  
\end{table}
Table~\ref{tab:error_frequency_models} provides model performance across different times of the day for the easy clock category. The results indicate that certain times pose greater challenges for MLLMs, with times such as 2:00, 4:00, and 5:00 being misclassified by all models. Errors on this simpler task highlight gaps in accurate clock reading.
\section{Metrics}
\label{app:metrics}
\subsection{ClockQA Metrics}
\label{app:metrics_clockqa}
We evaluate the clock-reading performance with the following metrics:
\paragraph{Exact Match (EM).}
The proportion of predictions that exactly match the ground truth time:
\begin{equation}
\text{Exact Match Accuracy} = \frac{1}{n} \sum_{i=1}^{n} \Bigl[ T_{\text{true},i} = T_{\text{pred},i} \Bigr].
\end{equation}
\paragraph{MAE (Seconds).}
The mean absolute error in seconds, with a 12-hour circular wraparound (\ie we measure the shorter way around the clock face with a maximum error of 21,600):
\begin{equation}
\text{MAE} = \frac{1}{n}\sum_{i=1}^{n} \min\Bigl(\bigl|T_{\text{true},i} - T_{\text{pred},i}\bigr|,;43200 - \bigl|T_{\text{true},i} - T_{\text{pred},i}\bigr|\Bigr).
\end{equation}
\paragraph{Hour and Minute Errors.}
Average absolute differences for hour and minute hands, each with a circular wraparound:
\begin{align}
    \text{MAE}_{\text{hours}} &= \frac{1}{n}\sum_{i=1}^{n} \min\Bigl(|H_{\text{true},i} - H_{\text{pred},i}|,\,12 - |H_{\text{true},i} - H_{\text{pred},i}|\Bigr), \\
    \text{MAE}_{\text{minutes}} &= \frac{1}{n}\sum_{i=1}^{n} \min\Bigl(|M_{\text{true},i} - M_{\text{pred},i}|,\,60 - |M_{\text{true},i} - M_{\text{pred},i}|\Bigr),
\end{align}
\subsection{CalendarQA Metrics}
\label{app:metrics_calendarqa}
For calendar-based reasoning, we employ standard classification metrics:

\paragraph{Accuracy (Acc).}
The fraction of correct predictions for the day of the week.

\paragraph{Precision (P), Recall (R), \& F1.}
Macro-averaged scores across different date categories.
\section{Experiment Details and Prompt Template}
\label{sec:exp_prompt_template}
For closed-source models, we used the default settings specified by their respective platforms. Open-source models were evaluated with a beam size of 4 and greedy sampling to ensure reproducibility. We used the following prompts for the models and also applied standard model-specific chat templates when available.
\begin{figure*}[htpb]
\centering
\begin{tcolorbox}[
  colback=gray!10!white,
  colframe=black!50!black,
  title=Prompt Template for ClockQA,
  fonttitle=\bfseries,
  halign title=flush center,
  width=1\textwidth
]
$<$Question$>$ + Only output the time.\\

Example: What time is shown on the clock in the given image? Only output the time.

\end{tcolorbox}
\vspace{-10pt}
\label{plan_clockqa_question_generation}
\end{figure*}
\begin{figure*}[htpb]
\centering
\begin{tcolorbox}[
  colback=gray!10!white,
  colframe=black!50!black,
  title=Prompt Template for CalendarQA,
  fonttitle=\bfseries,
  halign title=flush center,
  width=1\textwidth
]
$<$Question$>$ + in the given calendar image? Only output the day without explanation.\\

Example: Which day of the week is New Year's Day in the given calendar image? Only output the day without explanation.
\end{tcolorbox}
\vspace{-10pt}
\label{plan_calendarqa_question_generation}
\end{figure*}
\section{Sample of ClockQA and CalendarQA Predictions}
\label{sec:prediction_samples}
Sample images of clocks and calendars from the dataset along with model predictions.
\begin{table}[ht]
    \centering
    \begin{tabular}{c|c|c|c}
        \toprule
        \textbf{Image} & \textbf{Clock Type} & \textbf{Ground Truth} & \textbf{Model Predictions} \\
        \midrule
        \raisebox{-0.45\height}{\scalebox{0.8}{
        \includegraphics[width=0.2\textwidth]{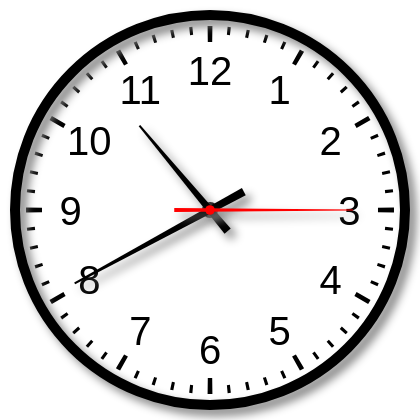}}} & Basic Dial & \textbf{10:40:15} &
        \begin{tabular}{l}
            GPT-4o: \textcolor{darkred}{8:22:15}\\
            Claude: \textcolor{darkred}{3:40:00}\\
            GPT-o1: \textcolor{darkred}{8:15:15}\\
            Gemini: \textcolor{darkred}{10:39:15}\\
            MiniCPM: \textcolor{darkred}{7:00:00}\\
            Qwen2: \textcolor{darkred}{10:10:10}\\
            Llama3: \textcolor{darkred}{10:03:30}\\
        \end{tabular}\\
        \midrule
          \raisebox{-0.45\height}{\scalebox{0.8}{
        \includegraphics[width=0.2\textwidth]{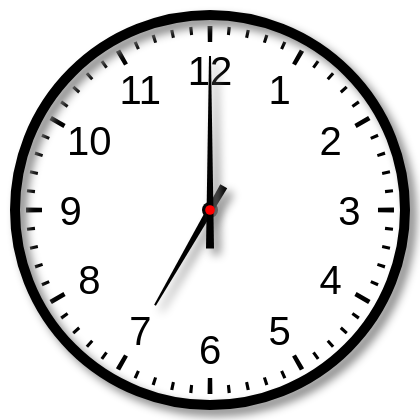}}} & Easy (Exact Hour) & \textbf{7:00:00} &
        \begin{tabular}{l}
            GPT-4o: \textcolor{darkred}{10:10:00}\\
            Claude: \textcolor{darkred}{11:58:30}\\
            GPT-o1: \textcolor{darkred}{12:35:00}\\
            Gemini: \textcolor{darkgreen}{7:00:00}\\
            MiniCPM: \textcolor{darkred}{12:00:00}\\
            Qwen2: \textcolor{darkred}{10:10:10}\\
            Llama3: \textcolor{darkred}{12:00:00}\\
        \end{tabular}\\
        \midrule
         \raisebox{-0.45\height}{\scalebox{0.8}{
        \includegraphics[width=0.2\textwidth]{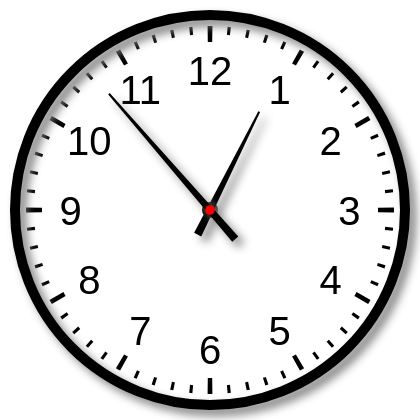}}} & No Second hand & \textbf{12:53:00} &
        \begin{tabular}{l}
            GPT-4o: \textcolor{darkred}{10:10}\\
            Claude: \textcolor{darkred}{10:10}\\
            GPT-o1: \textcolor{darkred}{10:10}\\
            Gemini: \textcolor{darkred}{10:59}\\
            MiniCPM: \textcolor{darkred}{10:09}\\
            Qwen2: \textcolor{darkred}{10:10}\\
            Llama3: \textcolor{darkred}{12:00}\\
        \end{tabular} \\
        \midrule
         \raisebox{-0.45\height}{\scalebox{0.8}{
        \includegraphics[width=0.2\textwidth]{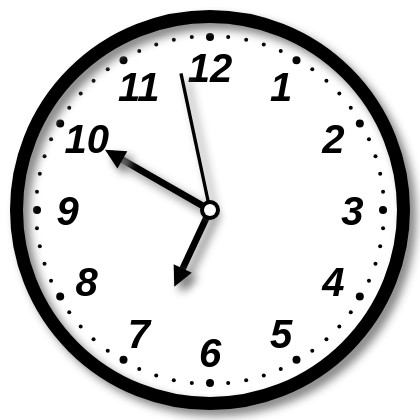}}} & Arrow hand & \textbf{6:49:58} &
        \begin{tabular}{l}
            GPT-4o: \textcolor{darkred}{10:08:46}\\
            Claude: \textcolor{darkred}{10:35:00}\\
            GPT-o1: \textcolor{darkred}{9:50:00}\\
            Gemini: \textcolor{darkred}{10:30:00}\\
            MiniCPM: \textcolor{darkred}{10:00}\\
            Qwen2: \textcolor{darkred}{11:11:11}\\
            Llama3: \textcolor{darkred}{10:10:00}\\
        \end{tabular} \\

        \midrule
          \raisebox{-0.45\height}{\scalebox{0.8}{
        \includegraphics[width=0.2\textwidth]{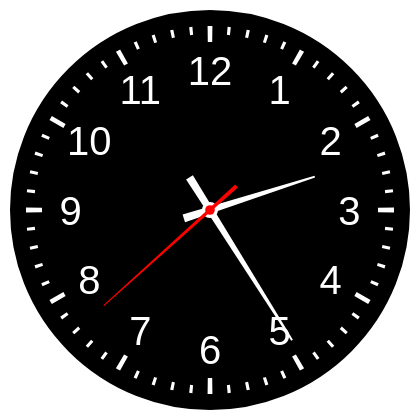}}} & Black Dial & \textbf{2:24:38} &
        \begin{tabular}{l}
            GPT-4o: \textcolor{darkred}{11:31:45}\\
            Claude: \textcolor{darkred}{10:10:30}\\
            GPT-o1: \textcolor{darkred}{9:25:40}\\
            Gemini: \textcolor{darkred}{10:25:39}\\
            MiniCPM: \textcolor{darkred}{10:09}\\
            Qwen2: \textcolor{darkred}{10:10:10}\\
            Llama3: \textcolor{darkred}{11:30:00}\\
        \end{tabular} \\
                \midrule

        \raisebox{-0.45\height}{\scalebox{0.8}{   
        \includegraphics[width=0.2\textwidth]{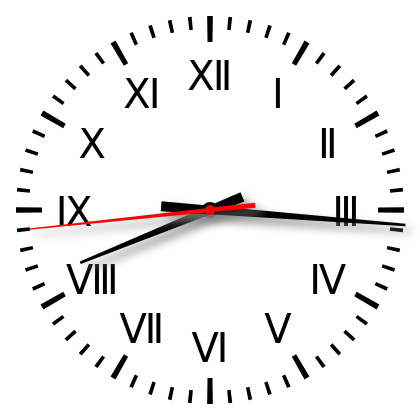}}} & Roman Numbers & \textbf{8:15:44} &
        \begin{tabular}{l}
            GPT-4o: \textcolor{darkred}{3:15:45}\\
            Claude: \textcolor{darkred}{8:40:30}\\
            GPT-o1: \textcolor{darkred}{10:15:45}\\
            Gemini: \textcolor{darkred}{2:42:25}\\
            MiniCPM: \textcolor{darkred}{10:09:09}\\
            Qwen2: \textcolor{darkred}{10:10:10}\\
            Llama3: \textcolor{darkred}{12:00:00}\\
        \end{tabular} \\
        \bottomrule
    \end{tabular}
    \caption{Clock image samples of different categories with model predictions.}
    \label{tab:model_predictions}
\end{table}
\begin{table}[t]
    \centering
     \renewcommand{\arraystretch}{1.2} 
    \begin{tabular}{c}
        \toprule
        \textbf{Calendar Image -- 2025} \\
        \midrule
        \includegraphics[width=0.8\textwidth]{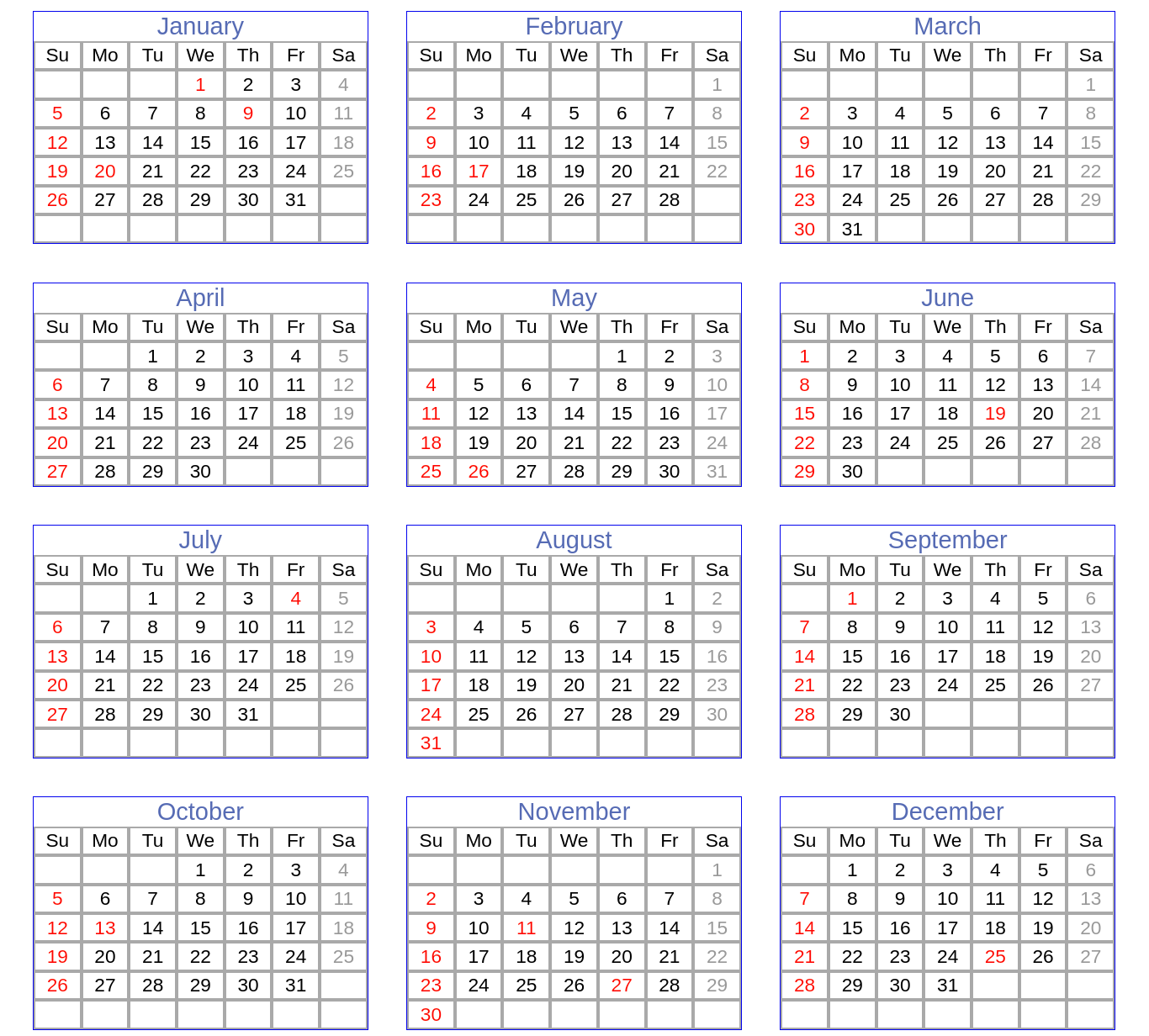} \\
    \end{tabular}
    \vspace{5pt}
    \begin{tabular}{p{0.95\textwidth}}
        \toprule
        \textbf{Question:} Which day of the week is New Year's Day in the given calendar image? \\
        \midrule
        \textbf{Ground Truth:} Wednesday \\
        \midrule
        \textbf{Model Predictions:}\\
            GPT-4o: \textcolor{darkgreen}{Wednesday}\\
            Claude: \textcolor{darkgreen}{Wednesday}\\
            GPT-o1: \textcolor{darkgreen}{Wednesday}\\
            Gemini: \textcolor{darkred}{Tuesday}\\
            MiniCPM: \textcolor{darkred}{Monday}\\
            Qwen2: \textcolor{darkred}{Monday}\\
            Llama3: \textcolor{darkred}{Sunday}\\

    \end{tabular}    
    \begin{tabular}{p{0.95\textwidth}}
        \toprule
        \textbf{Question:} Which weekday corresponds to the 100th day of the year (Assume January 1st is day 1.) in the given calendar image? \\
        \midrule
        \textbf{Ground Truth:} Thursday \\
        \midrule
        \textbf{Model Predictions:}\\
            GPT-4o: \textcolor{darkred}{Tuesday}\\
            Claude: \textcolor{darkred}{Monday}\\
            GPT-o1: \textcolor{darkgreen}{Thursday}\\
            Gemini: \textcolor{darkred}{Tuesday}\\
            MiniCPM: \textcolor{darkgreen}{Thursday}\\
            Qwen2: \textcolor{darkred}{Monday}\\
            Llama3: \textcolor{darkred}{Monday}\\
        \bottomrule
    \end{tabular}    
    \caption{Sample calendar image of the year 2025 with model predictions.}
    \label{tab:calendar_predictions_2025}
\end{table}
\begin{table}[t]
    \centering
     \renewcommand{\arraystretch}{1.2} 
    \begin{tabular}{c}
        \toprule
        \textbf{Calendar Image -- 2020} \\
        \midrule
        \includegraphics[width=0.8\textwidth]{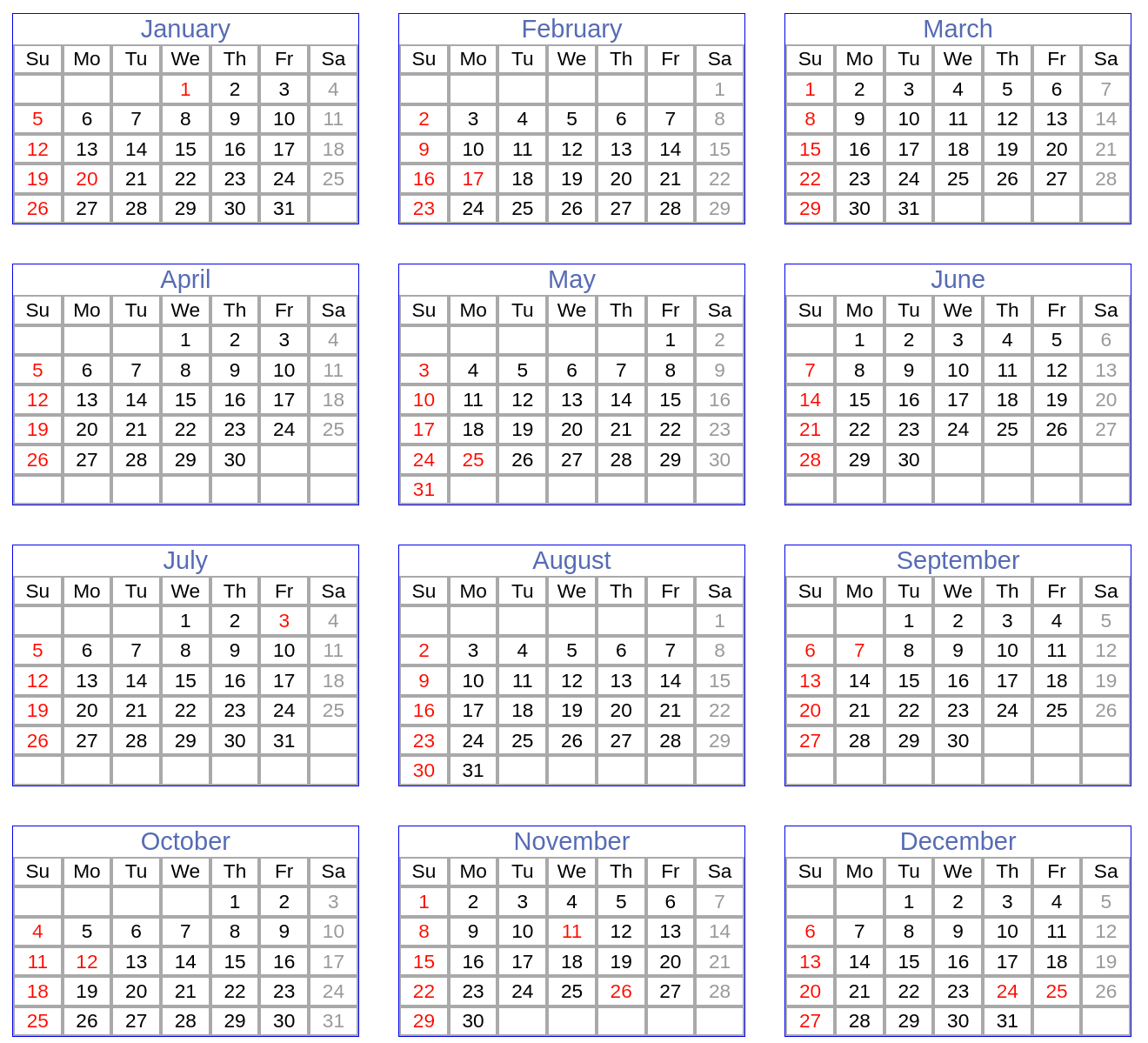} \\
    \end{tabular}
    \vspace{5pt}
    \begin{tabular}{p{0.95\textwidth}}
        \toprule
        \textbf{Question:} What's the day of the week on November 21st in the given calendar image? \\
        \midrule
        \textbf{Ground Truth:} Saturday \\
        \midrule
        \textbf{Model Predictions:}\\
            GPT-4o: \textcolor{darkred}{Friday}\\
            Claude: \textcolor{darkred}{Tuesday}\\
            GPT-o1: \textcolor{darkgreen}{Saturday}\\
            Gemini: \textcolor{darkred}{Thursday}\\
            MiniCPM: \textcolor{darkred}{Friday}\\
            Qwen2:  \textcolor{darkred}{Wednesday}\\
            Llama3: \textcolor{darkred}{Wednesday}\\

    \end{tabular}    
    \begin{tabular}{p{0.95\textwidth}}
        \toprule
        \textbf{Question:} Which day of the week is Christmas in the given calendar image? \\
        \midrule
        \textbf{Ground Truth:} Friday \\
        \midrule
        \textbf{Model Predictions:}\\
            GPT-4o: \textcolor{darkred}{Thursday}\\
            Claude: \textcolor{darkgreen}{Friday}\\
            GPT-o1: \textcolor{darkgreen}{Friday}\\
            Gemini: \textcolor{darkred}{Sunday}\\
            MiniCPM: \textcolor{darkred}{Sunday}\\
            Qwen2: \textcolor{darkred}{Monday}\\
            Llama3: \textcolor{darkred}{December 25}\\
        \bottomrule
    \end{tabular}    
    \caption{Sample calendar image of the year 2019 with model predictions.}
    \label{tab:calendar_predictions_2019}
\end{table}
\end{document}